\documentclass[french]{article}

\usepackage[T1]{fontenc}
\usepackage[utf8]{inputenc}
\usepackage{url}

\usepackage[top=2cm, bottom=2cm, left=3cm, right=3.5cm]{geometry}
\usepackage{hyperref}

\usepackage{graphicx}
\newcommand{\citep}[1]{\cite{#1}}
\newcommand{\fup}[1]{$^{#1}$}

\RequirePackage{graphicx}
\usepackage{booktabs} 
\usepackage{arydshln}
\usepackage{url}
\usepackage{hyperref}

\usepackage{xfrac} 
\RequirePackage{amsthm,amsmath,amsfonts,amssymb}

\newtheorem{definition}{Définition}

\newtheorem{question}{Question}

\usepackage{bm}

\usepackage{xspace}

\newcommand{\espace}{\vspace{10pt}}
\newcommand{\seq}[1]{\bm{#1}}

\usepackage{relsize}

\usepackage{pifont}

\usepackage{calc}

\usepackage{mathtools}

\newcommand*\gennoninclrel{\not\subseteq_{{\mathlarger{*}}}}
\newcommand*\partialnoninclrel{\not\subseteq_{G}}
\newcommand*\totalnoninclrel{\not\subseteq_{D}}

\usepackage{tikz}

\usepackage{setspace}

\title{\textbf{A survey on the semantics of sequential patterns with negation}}

\author{Thomas Guyet\fup{1}\\[6pt]
\fup{1} Inria -- Centre de Lyon, \href{https://team.inria.fr/aistrosight/}{AIstroSight}}

\date{thomas.guyet@inria.fr}

\begin{document}

\maketitle

\paragraph*{Abstract}
A sequential pattern with negation, or negative sequential pattern~\cite{Wang2019}, takes the form of a sequential pattern for which the negation symbol ($\neg$) may be used in front of some of the pattern's itemsets. 
Intuitively, such a pattern occurs in a sequence if negated itemsets are \textit{absent} in the sequence. 
Recent work~\cite{besnard2020semantics} has shown that different semantics can be attributed to these pattern forms, and that state-of-the-art algorithms do not extract the same sets of patterns.
This raises the important question of the interpretability of sequential pattern with negation. 

In this study, our focus is on exploring how potential users perceive negation in sequential patterns. Our aim is to determine whether specific semantics are more "intuitive" than others and whether these align with the semantics employed by one or more state-of-the-art algorithms. To achieve this, we designed a questionnaire to reveal the semantics' intuition of each user. 
This article presents both the design of the questionnaire and an in-depth analysis of the 124 responses obtained.

The outcomes indicate that two of the semantics are predominantly intuitive; however, neither of them aligns with the semantics of the primary state-of-the-art algorithms. As a result, we provide recommendations to account for this disparity in the conclusions drawn.


\section{Introduction}

Sequential pattern extraction is a classic class of data mining methods. Its objective is to extract subsequences (patterns) that frequently appear from a large dataset of sequences.
A pattern is considered frequent when it appears in at least $\sigma$ sequences, where $\sigma$ is user-defined.
For instance, consider the pattern $\langle e\ (ca)\ d\rangle$, which indicates that ``item $e$ is followed by the itemset $ca$ and then by item $d$ simultaneously''. In the table below, this pattern appears in 4 sequences ($\seq{p_0}$, $\seq{p_2}$, $\seq{p_3}$, and $\seq{p_4}$).

\begin{table}[h]
\caption{Example of a dataset containing five sequences over an alphabet of six items $\Sigma=\{a,b,c,d,e,f\}$. }\label{tab:intro}
\centering
\begin{tabular}{ll}
\hline
\textit{id} & \textit{Sequence}\\\hline
$\seq{p_0}$ & $\langle e\ (caf)\ d\ b\ e\ d\rangle$\\
$\seq{p_1}$ & $\langle c\ a\ d\ b\ e\ d\rangle$ \\
$\seq{p_2}$ & $\langle e\ (ca)\ d\rangle$ \\
$\seq{p_3}$ & $\langle d\ e\ (ca)\ b\ d\ b\ e\ f\rangle$ \\
$\seq{p_4}$ & $\langle c\ e\ b\ (fac)\ d\ e\ c\rangle$ \\\hline
\end{tabular}
\end{table}

These frequent patterns can be efficiently enumerated thanks to the anti-monotonicity property of the support measure (i.e., the number of occurrences of a pattern). Intuitively, the support of a pattern decreases with the pattern's size. This property, utilized by most algorithms in the literature, prevents enumerating patterns that are larger than those known a priori not to be frequent. This trick ensures the complete exploration of the search space while maintaining algorithm efficient.

Several studies \cite{cao2016nsp, guyet2020negpspan} have expanded the domain of sequential patterns by incorporating information about the absence of item occurrences. Such patterns are termed ``sequential patterns \textit{with negation}'' or ``\textit{negative sequential patterns}''. Sequential patterns with negation take the form of sequential patterns in which negation symbols, $\neg$, preceding certain items. The negation symbol indicates that the specified item must be absent from a sequence  for the pattern to be considered to occur. Intuitively, the pattern $\langle a\ \neg b\ c\rangle$ is recognized in a sequence if the latter contains an $a$ followed by a $c$, and $b$ is not present between the occurrences of $a$ and $c$. 

We advocate for a broader use of sequential patterns with negation in the process of mining datasets of sequences.
This type of pattern holds particular significance for data analysts, as it has the potential to unveil meaningful insights from the absence of events. 
For instance, in the context of health, the non-administration of a certain drug ($d$) might trigger an illness ($i$). 
When analyzing a database using a conventional sequential pattern mining algorithm, frequent patterns might indicate an illness occurrence without other co-occurring events. However, in the conventional semantics of sequential patterns, the absence of other events related to illness cannot be concluded from this pattern.  
Sequential patterns with negation, such as $\langle \neg d\ i\rangle$, bring to light the frequent co-occurrence of drug absence and the occurrence of an illness.

In this study, we would like to highlight possible interpretability issues of sequential patterns with negation. Indeed, Besnard and Guyet~\cite{besnard2020semantics} have demonstrated the existence of multiple semantics for these patterns. have demonstrated the existence of multiple semantics for these patterns, there is a risk of misinterpretation of extracted patterns in case the user and the algorithms do not share the same semantics. 
This concern is not solely theoretical; it manifests practically since the two state-of-the-art algorithms, eNSP~\cite{cao2016nsp} and {\sc NegPSpan}~\cite{guyet2020negpspan}, do not have the same semantics for the negation symbol~\cite{besnard2020semantics}. 
As a result, the patterns extracted by each of these algorithms need to be interpreted differently by the user.

Considering that a user does not necessarily seek to understand the intricacies of these patterns, we believe that the designers of pattern mining algorithms have to take care of possible misinterpretations of the outputs of their algorithms. 
Therefore, it is crucial to identify any possible disparity between the semantics used in an algorithm and the one that is perceived ``intuitively'' by users. 

In this article, we have therefore investigate three questions: 
\begin{enumerate}
\item Is there an ``intuitive'' semantics for patterns with negation? 
\item Does the ``intuitive'' semantics correspond to those actually employed by any of the algorithms?
\item What recommendations can be made regarding the use of patterns with negations?
\end{enumerate}

To address these questions,  our methodology involved designing a questionnaire to uncover the intuitive semantics of potential users of pattern mining algorithms. 
The details of the methodology of this survey are described in Section~\ref{sec:enquete}. 
Section~\ref{sec:semantics} presents the questions posed to users and makes explicit the potential alternative interpretations. 
The collected results from 124 participants are presented and analyzed in Section~\ref{sec:results}. 
We begin by introducing a brief overview of state-of-the-art algorithms for extracting sequential pattern with negations.

\section{State-of-the-art in sequential pattern extraction with negations}
The first endeavor in negative pattern extraction was presented by Savasere et al.~\cite{Savasere1998} in the context of itemset mining. Initial efforts toward sequential patterns with negation were made by Wu et al.~\cite{wu2004efficient} for association rules. Over time, several recent approaches have emerged to capitalize on advancements in pattern extraction techniques.

The eNSP algorithm extracts negative patterns by leveraging set operations between sets of sequences matched by frequent sequential patterns~\citep{cao2016nsp}. This approach circumvents the direct enumeration of patterns with negation that leads to efficient algorithms. Since then, numerous alternatives to this algorithm have been proposed, focusing on item utility~\cite{xu2017HighUtilityNSP}, repetitions~\cite{dong2018rnsp}, multiple support constraints~\cite{xu2017msnsp}, and more. Nonetheless, these methods do not rely on an antimonotonicity property of the support measure and they do not guarantee to extract all frequent patterns.

An alternative to eNSP is {\sc NegPSpan}~\cite{guyet2020negpspan}, which employs a distinct pattern semantics to harness the antimonotonicity property. This enables efficient and complete extraction following conventional pattern mining principles. The completeness of the mining process makes the approach more reliable as it guarantees to the user to not miss interesting patterns. And the implementation benefits from decades of pattern mining research to maintain the efficiency.

More recently, Wang et al.~\cite{Wang2021vm} introduced VM-NSP, an algorithm utilizing a vertical representation to enhance efficiency. For a comprehensive overview of recent developments in mining sequential pattern with negation, interested readers can refer to the work of Wang et al.~\citep{Wang2019}.

\medskip

In the initial stages, early approaches were compared without employing uniform pattern semantics. However, the recognition of distinct semantics has contributed to the clarification of the domain~\cite{besnard2020semantics}. Specifically, eight semantics of patterns with negations have been delineated. These eight variations stem from different interpretations of the notion of non-inclusion, occurrence, and inclusion relation. These notions, detailed in Section~\ref{sec:semantics}, have informed the design of our questionnaire.

\section{Survey on the Perception of Sequential Patterns with Negations}\label{sec:enquete}

\begin{figure}[tb]
\centering
\includegraphics[width=\textwidth]{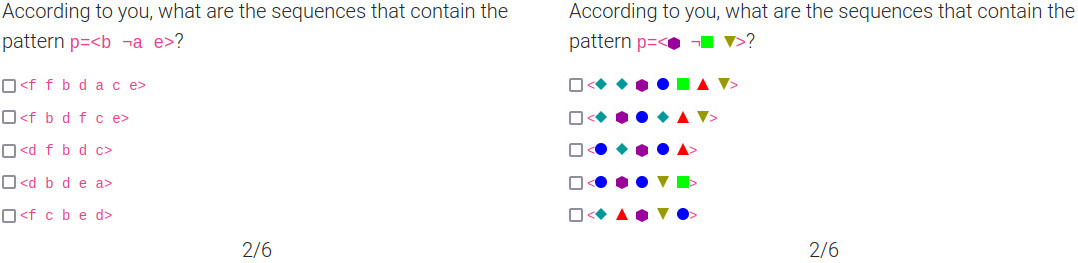}
\caption{Illustration of the two versions of the questionnaire: on the left, the classical view employing mathematical notations; on the right, the version employing colored shapes tailored for non-expert users. The use of colors and shapes provides redundancy while also catering to color-blind individuals.}
\label{fig:twoversions}
\end{figure}

The survey aims to identify the most intuitive semantics of sequential patterns with negation. The questionnaire is organized into three parts:
\begin{enumerate}
\item Evaluation of background knowledge in the domains of pattern mining and logic. In this part, participants are asked whether they are familiar with the concepts of pattern mining and whether they are computer scientists, logicians, or researchers. This information helps characterize potential biases within the participant group.
\item Verification of the understanding of sequential patterns (without negation) and the scope of negations. The general framework for the semantics of negative sequential patterns~\cite{besnard2020semantics} makes assumptions about the definition of a classical sequential pattern and the intuitive scope of the negation. Two questions assess whether participants adhere to these definitions. Correct answers are required for inclusion in the survey analysis.
\item Identification of the intuitive semantics of sequential patterns with negation. This third part constitutes the core of the questionnaire. Participants are asked to determine which sequences they believe contain a given pattern (see example in Figure~\ref{fig:twoversions}). The questions have been designed to unveil the semantics assumed by each participant. Thus, each participant is assigned one of the eight possible semantics. We refer to this questionnaire as revealing the intuitive semantics of participants, as they are not explicitly asked to state their preferred interpretations, but their interpretations are indirectly inferred from their answers.
\end{enumerate}

The questionnaire was distributed between December 2021 to March 2023. We used research mailing lists and non-research channels to collect responses from both experts and non-experts. The questionnaire is accessible via a standard web browser\footnote{\url{http://people.irisa.fr/Thomas.Guyet/negativepatterns/Survey/index.php}}. The questionnaire begins with explanations of sequential pattern concepts. It is designed to accommodate users with varying levels of mathematical comprehension by offering two versions: one employing letter notations and the other employing colored symbols. Figure~\ref{fig:twoversions} depicts the two alternative format to presenting a question.

The questionnaire is entirely anonymous, and the collected data only include dates and  answers to the questions.

\section{General Framework}

We now introduce the syntax of sequential patterns with negation which restricts the general definition of sequential patterns with negation to the ones introduced by Besnard and Guyet~\cite{besnard2020semantics}. In the following, let $[n] = {1, \dots, n}$ denote the set of the first $n$ integers, and let $\mathcal{I}$ denote a set of items (alphabet).

A subset $A = {a_1, a_2, \dots, a_m} \subseteq \mathcal{I}$ is called an \emph{itemset}. A \textit{sequence} $\seq{s}$ is of the form $\seq{s} = \langle s_1, s_2, \dots, s_n \rangle$, where $s_i$ is an itemset.

\begin{definition}[Sequential pattern with negation]\label{def:negativepattern}
A sequential pattern with negation $\seq{p} = \langle p_1, \neg q_1, \-p_2, \-\neg q_2, \-\dots, p_{n-1}, \neg q_{n-1}, p_n \rangle$ is such that $p_i \in 2^{\mathcal{I}} \setminus \emptyset$ for all $i \in [n]$ and $q_i \in 2^{\mathcal{I}}$ for all $i \in [n-1]$. $\seq{p}^+ = \langle p_1, p_2, \dots, p_n \rangle$ denotes the \textit{positive part} of $\seq{p}$.
\end{definition}

The semantics of patterns relies on the containment relation, which specifies how to determine whether a pattern occurs (is contained) or not in a sequence. This relation utilizes the notion of occurrence of a (positive) sequential pattern in a sequence, formally defined as follows:

\begin{definition}[Occurrence of a sequential pattern]\label{def:positivepattern_embedding}
Let $\seq{s} = \langle s_1, s_2, \dots, s_n \rangle$ be a sequence and $\seq{p} = \langle p_1, p_2, \dots, p_m \rangle$ be a sequential pattern, $\seq{e} = (e_i)_{i \in [m]} \in [n]^m$ is an \emph{occurrence} of the pattern $\seq{p}$ in the sequence $\seq{s}$ if $p_i \subseteq s_{e_i}$ for all $i \in [m]$ and $e_{i} < e_{i+1}$ for all $i \in [m-1]$.
\end{definition}

The understanding of this definition (explained at the beginning of the questionnaire) is verified through the following question.

\begin{question}[Occurrence of a sequential pattern]\label{ex:positive}
Let $\seq{p} = \langle (ca)\ d\ e \rangle$ be a sequential pattern, indicate in which sequences of Table \ref{tab:intro} the pattern $\seq{p}$ occurs.
\end{question}

The expected answers to this question are the sequences $\seq{p_0}$, $\seq{p_3}$, and possibly $\seq{p_4}$. Sequence $\seq{p_0}$ allows us to verify the understanding that $(ca)$ appears in $(caf)$ as per our definitions. Sequence $\seq{p_1}$ verifies that all the elements of $(ca)$ appear together (and not just a subset). Sequence $\seq{p_2}$ allows us to verify the understanding of the importance of the occurrence order in the sequence. Sequence $\seq{p_3}$ lets us verify the understanding of the notion of a \textit{gap}: it is possible to have itemsets in the middle of an occurrence (e.g., the occurrence of $b$ between $d$ and $e$). Lastly, the final sequence presents an itemset whose items are not ordered. If $\seq{p_4}$ is not deemed to contain $\seq{p}$, it would indicate a user's sensitivity to the order within an itemset (which is classically not the case).

Likewise, the semantics of sequential patterns with negation are based on a containment relation. A pattern with negation, $\seq{p}$, is contained in a sequence $\seq{s}$ if $\seq{s}$ contains a subsequence $\seq{s}'$ such that each positive set of $\seq{p}$ (denoted as $p_i$) is included in an itemset of $\seq{s}'$ (in proper order), and all the negation constraints expressed by $\neg q_i$ are also satisfied. The negation constraint on $q_i$ then applies to the subsequence of $\seq{s}'$ located between the occurrence of the positive itemset preceding $\neg q_i$ in $\seq{p}$ and the occurrence of the positive itemset following $\neg q_i$ in $\seq{p}$.

This definition determines the scope of the negation, which is specific to the framework we are working in. Ensuring that users share this definition is paramount. The subsequent question enables us to affirm this understanding.

\begin{question}[Scope of the negation]\label{ex:portee}
Let $\seq{p}=\langle c\neg d\ e \rangle$ be a pattern with negation, indicate the sequences of the table below in which, according to you, $\seq{p}$ occurs.

\centering
\begin{tabular}{ll}
\hline
\textit{id} & \textit{Sequence}\\\hline
$\seq{s_0}$ & $\langle f\ f\ c\ b\ d\ a\ e\rangle$ \\
$\seq{s_1}$ & $\langle f\ c\ b\ f\ a\ e\rangle$ \\
$\seq{s_2}$ & $\langle b\ f\ c\ b\ a\rangle$ \\
$\seq{s_3}$ & $\langle b\ c\ b\ e\ d\rangle$ \\
$\seq{s_4}$ & $\langle f\ a\ c\ e\ b\rangle$ \\\hline
\end{tabular}
\end{question}

In this question, it seems reasonable to consider that $\seq{p}$ occurs in $\seq{s_1}$, $\seq{s_3}$ (since $d$ is outside the assumed scope of the negation), and $\seq{s_4}$. Participants who do not tick $\seq{s_4}$ likely interpret the constraint $\neg d$ as referring to the occurrence of an element other than $d$ (which is not consistent with the definitions proposed above). If $\seq{p_0}$ is deemed to contain $\seq{p}$, it is likely that the constraint $\neg d$ is understood to strictly follow $c$ which is not a situation considered in our framework.

\section{Questions on the semantics of negations}
\label{sec:semantics}
In this section, we take up the questions of the third part of the questionnaire and we explain the different interpretations revealed by the answers given by the participants.
There are three questions. Each question is dedicated to one dimension of the semantics of negative sequential pattern, and they cover all dimensions identified in~\cite{besnard2020semantics}.

\subsection{Itemset non-inclusion}
\begin{question}\label{ex:non-inclusion}
Let $\seq{p}=\langle d\ \neg (af)\ b \rangle$ be a sequential pattern with negation, indicate the sequences of the table below in which, according to you, $\seq{p}$ occurs.

\centering
\begin{tabular}{ll}
\hline
\textit{id} & \textit{Sequence}\\\hline
$\seq{i_0}$ & $\langle e\ e\ d\ a\ b\ e\rangle$  \\
$\seq{i_1}$ & $\langle d\ (af) b\ c\rangle$  \\
$\seq{i_2}$ & $\langle e\ d\ (fc)\ b\rangle$  \\
$\seq{i_3}$ & $\langle e\ c\ d\ (ec)\ b\rangle$ \\
$\seq{i_4}$ & $\langle d\ (fa)\ b\ e\rangle$ \\\hline
\end{tabular}
\end{question}

This question is designed to unveil the interpretation of the inclusion relation between itemsets. Each sequence in the table contains the positive part of the pattern, $\seq{p}^+=\langle d\ b \rangle$, with only one itemset between the occurrences of $d$ and $b$. These sequences prompt inquiry into the non-inclusion of the $(af)$ itemset in $a$, $(af)$, $(fc)$, $(ec)$, or $(fa)$. If a participant ticks the sequences $i_0$, $i_2$, and $i_3$, we can deduce that they regard the presence of at least one element of the itemset $(af)$ to ``validate'' the negation. This is referred to as ``partial non-inclusion''. On the other hand, if only sequence $\seq{i_3}$ is ticked, it suggests that the participant considers that all items in the itemset must be present to ``validate'' the negation. This is referred to as ``total non-inclusion''. Additionally, sequence $\seq{i_4}$ is included to examine whether the order of items in the itemset matters to participants and whether their response aligns with their answer to sequence $\seq{p_4}$ in Question~\ref{ex:positive}.

More formally, this question discriminates between two choices of inclusion between two itemsets, $P$ and $I$:
\begin{itemize}
\item Partial non-inclusion: $P\partialnoninclrel I \Leftrightarrow \exists e \in P$, $e \notin I$
\item Total non-inclusion: $P\totalnoninclrel I \Leftrightarrow \forall e \in P, e \notin I$
\end{itemize}

Partial non-inclusion means that $P \setminus I$ is non-empty, while total non-inclusion means that $P$ and $I$ are disjoint. In the following, the symbol $\gennoninclrel$ denotes a relation of non-inclusion between itemsets, either $\partialnoninclrel$ or $\totalnoninclrel$.

\subsection{Embedding of a pattern with negation}

\begin{question}[Embedding of a pattern with negation]\label{ex:embedding}
Let $\seq{p}=\langle f\ \neg (ea)\ d \rangle$ be the sequential pattern with negation, indicate the sequences from the table below in which, according to you, $\seq{p}$ occurs.

\centering
\begin{tabular}{ll}
\hline
\textit{id} & \textit{Sequence}\\\hline
$\seq{e_0}$ & $\langle b\ b\ f\ c\ e\ d\ b\rangle$ \\
$\seq{e_1}$ & $\langle b\ f\ e\ a\ c\ b\ d\rangle$ \\
$\seq{e_2}$ & $\langle f\ c\ (ea)\ b\ c\ d\ c\rangle$ \\
$\seq{e_3}$ & $\langle b\ c\ f\ b\ c\ c\ d\rangle$ \\\hline
\end{tabular}
\end{question}

The form of the pattern $\seq{p}=\langle f\ \neg (ea)\ d \rangle$ mirrors that of the previous question, differing by a permutation of letters. Each sequence in the table contains the positive part of $\seq{p}$, i.e. $\langle f\ d\rangle$. 
The primary difference is that there are multiple itemsets between the occurrences of $f$ and $d$. 
Participants must decide which itemset(s) of the sequence to compare with the negated itemsets of the pattern.

First and foremost, we anticipate participants to deduce that $\seq{p}$ occurs in $\seq{e_3}$ (there is clearly neither $e$ nor $a$ here) but that $\seq{p}$ does not occur in $\seq{e_2}$ (the itemset $(ea)$ is found in the scope of the negation). 
The sequence that unveil the participant semantics is $\seq{e_1}$. Notably, this sequence comprises both elements of the negated itemset ($e$ and $a$), but in two separated itemsets of the sequence. 
The participant who does not tick it (i.e. he/she considers that $e$ does not occur in $e_1$) uses the notion of ``soft-embedding'': $e$ and $a$ would have to appear together to ``validate'' the negation (as in the case of $e_2$).  
The participant who ticks it consider that the negation constraint applies across the entire set of itemsets within the negation's scope. The interpretation is termed \textit{strict-embedding}.

Furthermore, $\seq{e_0}$ unveils the notion of non-inclusion discussed earlier: in the case of partial non-inclusion, $\seq{p}$ occurs in~$\seq{e_0}$, but not if we consider a total non-inclusion. Thus, this sequence serves to assess the consistency of responses.

\espace

Two interpretations have been distinguished: strict- and soft-embeddings. They can be formally defined as follows:
Let a sequence $\seq{s}=\langle s_1,\dots\, s_n\rangle$ and a pattern with negation $\seq{p}=\langle p_1,\dots\ \neg q_1,\dots\ \neg q_{m-1}\ p_m\rangle$. 
We say that $\seq{e}=(e_i)_{i\in[m]}\in[n]^m$ is a soft-embedding of $\seq{p}$ in the sequence $\seq{s}$ iff:
\begin{itemize}
\item $p_i \subseteq s_{e_i}$ for all $i\in[m]$
\item $q_i \gennoninclrel s_j,\;\forall j\in [e_{i}+1,e_{i+1}-1]$ for all $i\in[m-1]$
\end{itemize}
We say that $\seq{e}=(e_i)_{i\in[m]}\in[n]^m$ is a strict-embedding of $\seq{p}$ in the sequence $\seq{s}$ iff:
\begin{itemize}
\item $p_i \subseteq s_{e_i}$ for all $i\in[m]$
\item $q_i \gennoninclrel \bigcup_{j\in [e_{i}+1,e_{i+1}-1]} s_j$ for all $i\in[m-1]$
\end{itemize}

Intuitively, the soft-embedding considers the non-inclusion of $q_i$ for each of the itemsets within the positional range $[e_{i}+1,e_{i+1}-1]$ while the strict-embedding considers the non-inclusion across the union of the itemsets at those same positions. 
The interval corresponds to the itemsets of the sequence that lie strictly between the occurrences of the itemsets surrounding $q_i$.

\subsection{Multiple occurrences}

\begin{question}[Multiple occurrences of a pattern with negation]\label{ex:occurrence}
Let $\seq{p}=\langle b\ \neg e\ f \rangle$ be a negative sequential pattern, indicate the sequences of the table below in which, according to you, $\seq{p}$ occurs.

\centering
\begin{tabular}{ll}
\hline
\textit{id} & \textit{Sequence} \\\hline
$\seq{o_0}$ & $\langle b\ a\ f\ d\ b\ d\ f\rangle$  \\
$\seq{o_1}$ & $\langle b\ a\ f\ d\ e\ b\ d\ f\rangle$  \\
$\seq{o_2}$ & $\langle d\ b\ e\ c\ a\ d\ f\ b\ d\ e\ f\rangle$  \\
$\seq{o_3}$ & $\langle b\ a\ f\ b\ a\ e\ f\rangle$ \\\hline
\end{tabular}
\end{question}

In this question, each sequence contains multiple occurrences of the positive part of the pattern, $p^+=\langle b\ f \rangle$. Notably, there are even non-nested occurrences of $\langle b\ f \rangle$ in each sequence to underscore this. 
Given that the negation constraint pertains only to the item $e$, whatever the choices of non-inclusion and embedding interpretations, the question centers on the interpretation of these multiple occurrences.
Two alternative interpretations are anticipated:
\begin{itemize}
\item The first interpretation considers that once an occurrence of the positive part, $\langle b\ f \rangle$, fulfills the negation constraint, the sequence contains the pattern. This is termed a ``weak occurrence''. 
Ticking sequences $\seq{o_0}$, $\seq{o_1}$, and $\seq{o_3}$ indicates alignment with this interpretation.
\item The second interpretation holds that if any occurrence of the positive part fails to satisfy the negation constraint, the sequence does not contain the pattern. This is termed a ``strong non-occurrence''. In Question~\ref{ex:occurrence}, participants subscribing to this view solely ticked  $\seq{o_0}$, as all other sequences possess at least one occurrence of $\langle b\ f \rangle$ with an interstitial $e$. 
However, sequence $\seq{o_1}$ might pose a challenge for those with this interpretation. It contains two minimal occurrences~\cite{mannila1997discovery} of $\langle b\ f \rangle$ that meet the negation constraint, alongside an occurrence involving the first~$b$ and the last~$f$ which does not satisfy the negation constraint.\footnote{In sequence mining, a minimal occurrence~\cite{mannila1997discovery} is an occurrence of a pattern whose extent within the sequence does not contain another occurrence of the same pattern. For instance, in the sequence $\langle {\color{red}b}\ {\color{blue}b\ f}\ {\color{red}f}\rangle$, the blue occurrence of $\langle b\ f\rangle$ is minimal, but not the red one.}  
This subtlety may be difficult to detect for those unfamiliar with sequences. Hence, it is advisable to assess the interpretation solely based on the absence of $\seq{o_3}$. 
When the participant ticks $\seq{o_1}$, we assign to him/her a specific attention to minimal occurrences.
\end{itemize}

\espace

Finally, the three dimensions of interpretation for negation combine to establish eight distinct semantics, each characterized by its containment relations as studied in~\cite{besnard2020semantics}. 
The three questions above were strategically crafted to individually delve into each of the three dimensions underlying the semantics of sequential patterns with negation.
Notably, this approach illustrates how the question design facilitates the assignment of a specific semantics to a participant based on their provided responses.

\section{Analysis of the questionnaire answers}\label{sec:results}
By the conclusion of the survey period, we had amassed 124 fully completed questionnaires. Participants' self-assessed expertise in pattern mining is distributed as follows: 40 novices (level 0), 54 with knowledge of data science (level 1), and 27 who identified themselves as familiar with pattern mining (level 3). In terms of background, 79 participants identified themselves as computer scientists, 82 as researchers, and 23 as logicians.

The average number of attempts made to comprehend the notion of pattern occurrence was $1.27\pm 0.49$, with attempts ranging from 1 to 5. Notably, 102 participants answered correctly on their initial attempt. It is worth noting that among the participants with knowledge of data analysis (out of 24), 6 requires more than one attempt to arrive at the correct answer.

The objective of the questionnaire analysis is to identify clusters of individuals who selected the same answers, i.e., who have the same intuitive semantics of sequential patterns with negation. 
This process unfolds in two stages:
\begin{enumerate}
\item Initially, we analyze the results question by question, focusing individually on each dimension of the semantics of negative sequential patterns
\item The analysis is then complemented by a global analysis of semantics.
\end{enumerate}

In the preceding section, we determined the expected responses for each question.
We propose the utilization of formal concept analysis (FCA) to achieve a comprehensive overview of the outcomes. FCA is a data analysis technique that identifies concepts within a dataset. Each concept is defined by its intention, which represents the set of selected answers, and its extension, which enumerates all individuals who select those answers.
These extracted concepts are ``closed'', meaning that their extension is maximal for their intention, and vice versa. FCA empowers us to succinctly represent the answers in a concept lattice. Through this lattice, we visualize all subgroups of individuals who provided identical answers. FCA has previously found application in questionnaire analysis~\citep{belohlavek2011evaluation}. For our practical implementation, we employed the GALACTIC tool~\citep{galactic} to construct our lattices.

\subsection{Analysis of each dimension of semantics}
\begin{figure}[tbh]
\centering
\includegraphics[width=.3\textwidth]{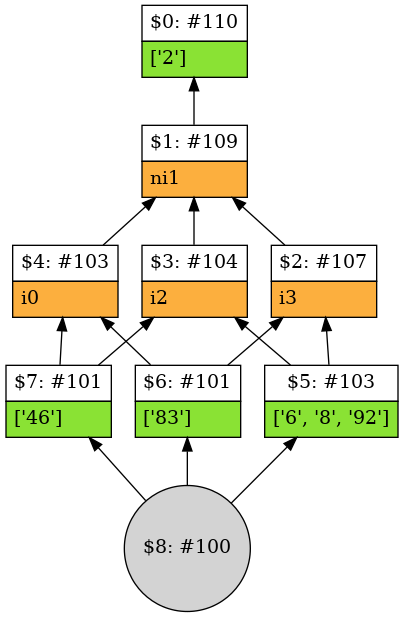}
\caption{Concepts extracted from the answers to Question~\ref{ex:non-inclusion}: non-inclusion of an itemset. 
Each concept is illustrated by a box containing different elements: the generators on an orange background (representing possible answers to the questions), and the prototypes on a green background. The size of the extension is indicated with a \texttt{\#}. Each concept indicates the intention as a set of ticked sequences (refer to the tables presented in the examples). 
In the responses to the questions, \texttt{i0} indicates that the participant ticked the sequence $\seq{i_0}$, and \texttt{ni1} (prefixed with \texttt{n}) indicates that the participant \textit{did not} tick the sequence $\seq{i_1}$.}
\label{fig:concepts_noninclusion}
\end{figure}

\begin{figure}[tbh]
\centering
\includegraphics[width=.5\textwidth]{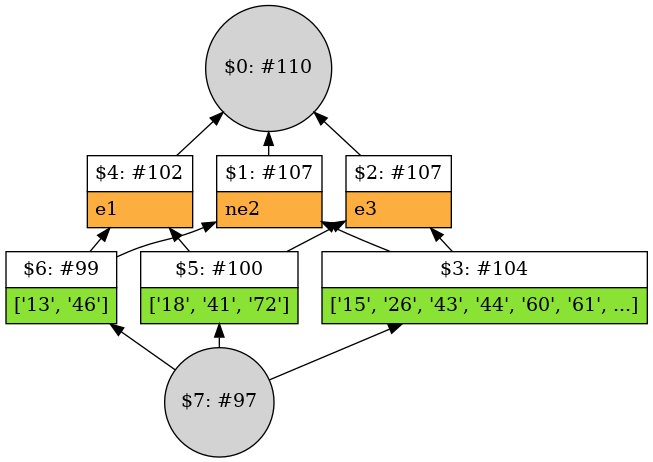}
\caption{Concepts extracted from responses to the Question~\ref{ex:embedding} relating to embeddings (see Figure~\ref{fig:concepts_noninclusion} for legend details).}
\label{fig:concepts_emb}
\end{figure}

\begin{figure}[tbh]
\centering
\includegraphics[width=.35\textwidth]{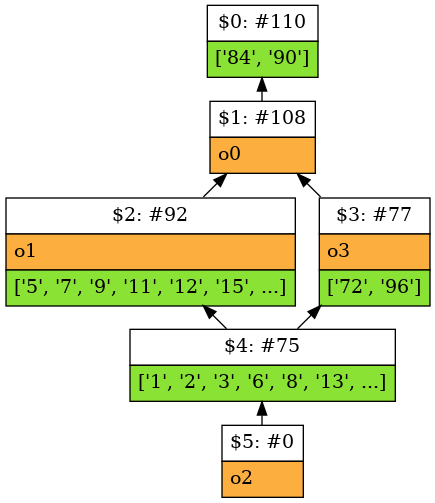}
\caption{Concepts extracted from responses to Question~\ref{ex:occurrence} relating to multiple occurrences (see Figure \ref{fig:concepts_noninclusion} for legend details).}
\label{fig:concepts_occ}
\end{figure}

In this section, we analyze the responses to questions~2 to~5. It should be noted that participants are required to answer Question 1 correctly to proceed with the questionnaire. As a result, the analysis of answers to this question may not be significant.
First, we focus on the answers to the question regarding the scope of negations. Subsequently, we delve into the analyze of the three dimensions of the semantics of patterns with negation: the non-inclusion of itemsets, embeddings, and multiple occurrences. Tables~\ref{tab:portee} to~\ref{tab:multocc} provide a synthetic account of each of the interpretations. Furthermore, Figures~\ref{fig:concepts_noninclusion} to~\ref{fig:concepts_occ} depict the concept lattices obtained for each of these questions to give a more global picture of the responses.

\begin{table}[th]
\caption{Results on the question of the scope of negation.}\label{tab:portee}
\centering
\begin{tabular}{lcc}
\hline
\textbf{Scope} & \textbf{Count} & \textbf{Percentage} \\ \hline
Conform & 101 & 81.4\%\\
Conform except $\seq{s_4}$ & 9 & 7.3\%\\
Alternative & 14 & 11.3\%\\
\hline
\end{tabular}
\end{table}

Regarding the scope of the negations, 101 participants provided answers that corresponded with the expected understanding of negation scope (see Table~\ref{tab:portee}). It is interesting to note that 9 people who selected $\seq{s_1}$ and~$\seq{s_3}$ did not select $\seq{s_4}$. This discrepancy suggests that, for them, negating an itemset means negating the event itself, rather than negating the presence of the event.\footnote{NB: In the following questions, all sequences have at least one ``neutral'' event where an itemset with negation is expected.} The remaining marginal differences (14 people) are assumed to be omissions or errors. As their grasp of the scope of negation might differ, these individuals were excluded from further results analysis, ensuring the interpretability of the responses. Therefore, the further analysis is based on 110 completed answers.

\begin{table}[th]
\caption{Responses to the question of non-inclusions (number and percentage).}\label{tab:non-incl}
\centering
\begin{tabular}{lcc}
\hline
\textbf{Interpretation} & \textbf{Count} & \textbf{Percentage} \\ \hline
Partial non-inclusion & 100 & 90.9\%  \\
Total non-inclusion & 3 & 2.7\% \\
Other & 7 & 6.4\% \\
\hline
\end{tabular}
\end{table}

Regarding the non-inclusion of itemsets (Table~\ref{tab:non-incl} and Figure~\ref{fig:concepts_noninclusion}), we can observe that the majority of participants (100) selected the response triple $\seq{i_0}$, $\seq{i_2}$, and $\seq{i_3}$ aligning with the interpretation of partial non-inclusion (concept \texttt{\$8} in Figure~\ref{fig:concepts_noninclusion}). Only 3 people considered the total non-inclusion interpretation. 
An interesting observation pertains to the 22 participants who considered that the sequence $\seq{i_4}$ contains the pattern. They believe that $(fa)$ is not incompatible with $(af)$. These participants spanned across varying levels of expertise: 8, 11, and 3, respectively, for levels 0, 1, and 2. Unsurprisingly, people knowledgeable in pattern mining (level 2) are, in proportion, less represented among people who are inclined to differentiate between~$(fa)$ and~$(af)$.


\begin{table}[th]
\caption{Responses to the question of embeddings.}\label{tab:emb}
\centering
\begin{tabular}{lcc}
\hline
\textbf{Interpretation} & \textbf{Count} & \textbf{Percentage} \\ \hline
Strict-embedding & 97 & 88.2\% \\
Soft-embedding & 7 & 6.3\% \\
Other & 6 & 5.5\% \\
\hline
\end{tabular}
\end{table}

Moving on to the analysis of the embeddings (Table~\ref{tab:emb} and Figure~\ref{fig:concepts_emb}), the sequence $\seq{e_1}$ allows us to distinguish the participants' intuition. For Table~\ref{tab:emb}, we also ensure that the answers are correct for $\seq{e_2}$ and $\seq{e_3}$; otherwise, we categorize the answer as ``other''. 
Once again, we observe a pronounced trend in the results. 97 participants subscribed to the soft-embedding interpretation (Concept \texttt{\$7} in Figure~\ref{fig:concepts_emb}). Concept~\texttt{\$3} corresponds to individuals who did not select $\seq{e_1}$, indicative of a strict-embedding interpretation.


\begin{table}[tbh]
\caption{Responses to the question of multiple occurrences.}\label{tab:multocc}
\centering
\begin{tabular}{lcc}
\hline
\textbf{Interpretation} & \textbf{Count} & \textbf{Percentage} \\ \hline
Weak occurrence & 75 & 69.2\% \\
Strong occurrence & 31 & 28.2\% \\
Other & 4 & 3.6\% \\
\hline
\end{tabular}
\end{table}

Lastly, regarding the analysis of the inclusion relations (Table~\ref{tab:multocc} and Figure~\ref{fig:concepts_occ}), two balanced groups of participants emerge. 75 participants have exclusively identified the three sequences corresponding to the notion of a weak occurrence. They are represented by Concept \texttt{\$3} of Figure~\ref{fig:concepts_occ}. On the other hand, 31 participants exclusively selected the sequence $\seq{o_0}$ (Concept~\texttt{\$1}). The latter group preferred the interpretation of a strong occurrence. 
Among these 31 participants, 15 did not select the $\seq{o_1}$ sequence, while 16 did (Concept~\texttt{\$2}). The latter group tends to align more with the notion of minimal occurrence.

\subsection{Global semantics analysis}
Questions~\ref{ex:non-inclusion} to~\ref{ex:occurrence} assign each participation to an interpretation of one of the three dimensions that constitute the semantics of a pattern with negation (according to the framework of Besnard and Guyet~\cite{besnard2020semantics}. We now investigate if there are dominant semantics (combinations of interpretation choices for the three dimensions) among the eight possibilities.

\begin{figure*}[tb]
\centering
\includegraphics[width=.8\textwidth]{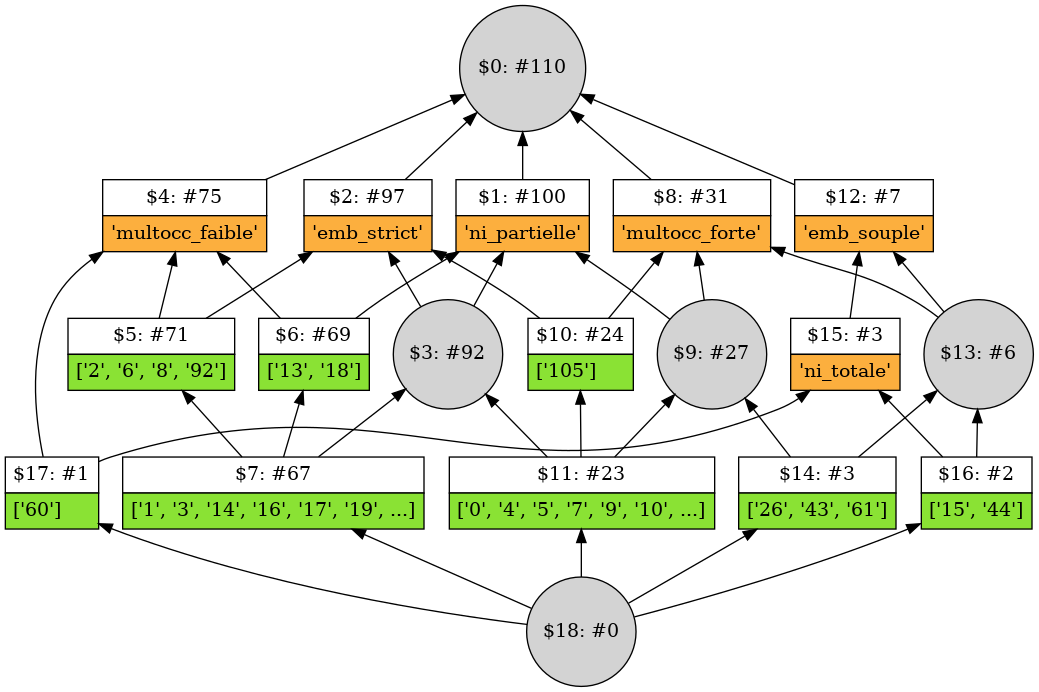}
\caption{Concepts extracted from the attributions made for each dimension.}
\label{fig:global}
\end{figure*}

Figure~\ref{fig:global} provides a summary of the survey responses. It presents the concept lattice that represents the semantics of patterns with negation. The five prototypes at the bottom level describe the five semantics (and their representation in the data) that the participants used. Among the 110 participants, 96 were assigned an intuitive semantic by the questionnaire. The remaining 14 participants had at least one question for which no clear interpretation was identified. These individuals are categorized in the intermediate concepts (prototypes \texttt{\$5}, \texttt{\$6}, and \texttt{\$10}; generator \texttt{\$15}; and concepts \texttt{\$3}, \texttt{\$9}, and \texttt{\$13}).

One noteworthy observation is that a significant proportion of participants can be attributed a semantic, suggesting that the same individuals likely provided ``alternative'' answers to different questions. This outcome reinforces the reliability of the collected answers.

Furthermore, the figure highlights the main finding of this study: there are two primary intuitively used semantics:
\begin{itemize}
\item the first is partial non-inclusion, with soft-embedding and strong-occurrences, accounting for 23.9\% of participants, and 
\item the second is partial non-inclusion, with soft-embedding and weak-occurrences, accounting for 69.8\%
\end{itemize}
The representation of the other semantics is marginal.

\espace

Additionally, we sought to compare the populations defined by their choice of semantics by analyzing their responses to profile questions. To do this, we conducted a statistical test to compare the distributions of expertise levels using Student's t-test. The results show no significant difference between the groups.
In conclusion, we find that the intuition of a semantics  is not inherently linked to a particular expertise in computer science or data science.

\section{Preferred semantics vs state-of-the-art algorithms}

As a preliminary summary, the analyses reveal the absence of a single shared semantic among participants, but rather the presence of two dominant semantics. These results prompt a comparison with the choices made by two prominent algorithms in the field:
\begin{itemize}
\item eNSP employs total non-inclusion, with soft-embedding and strong-occurrences
\item {\sc NegPSpan} employs total non-inclusion, with soft-embedding and weak-occurrences
\end{itemize}

Firstly, neither of the algorithms aligns with the participants' intuitive understanding, as both rely on total non-inclusion of itemsets, whereas partial non-inclusion appears to be the most intuitive. 
One possible explanation for this algorithmic choice is that partial non-inclusion is antimonotonic, while the total non-inclusion is monotonic. The latter is less straightforward to exploit algorithmically. Therefore, the most intuitive semantics may not be the most suitable from an algorithmic perspective.

In practice, this raises concerns about potential misinterpretation of patterns extracted by these state-of-the-art algorithms. Without explicitly defining their semantics, the results of this study indicate that the patterns will be interpreted differently from the intended interpretation used for their extraction. This poses a significant challenge for the practical use of these algorithms.

In light of these findings, several recommendations emerge:
\begin{enumerate}
\item \textbf{Singleton-only negations}: Consider limiting negations to singletons only. This adjustment would make partial and total non-inclusions equivalent, potentially reducing confusion and aligning better with participants' intuition.

\item \textbf{Algorithmic Adaptations}: Develop alternative algorithms tailored to the partial non-inclusion semantics. While these adaptations are algorithmically feasible, their computational performance should be rigorously compared to existing algorithms to assess their efficiency and competitiveness.
Given that NegPSpan adheres more closely with the intuition of a larger number of participants, consider favoring the extension and utilization of the NegPSpan algorithm.

\item \textbf{Distinct Syntaxes}: Promote the adoption of distinct syntaxes for each semantic interpretation. This approach can help differentiate and avoid confusion between different interpretations. This recommendation serves as a practical solution to address the challenges faced by the pattern mining community regarding sequential patterns with negations.
\end{enumerate}

While preferred semantics have been identified through our survey, we recognize that all semantics might have their uses depending on the data context. Resolving this challenge might involve designing algorithms capable of extracting various types of negative sequential patterns. This avenue has been explored in~\cite{besnard:hal-03025560} using a declarative pattern mining framework, although scalability to large datasets remains a limitation.

\section{Discussion}
In this part, we discuss the methodology  employed for conducting the survey. 
However, it's important to acknowledge several limitations associated with our approach.
Firstly, the survey encompassed only a limited number of questions that enabled a precise profiling of participants.  Consequently, our understanding of whether the surveyed population accurately represents potential users of pattern mining algorithms remains constrained. Additionally, the questionnaire was primarily disseminated through academic channels, which may introduce bias in the responses.

A second limitation of the questionnaire is the lack of redundancy in the questions. Each dimension of the semantics of patterns with negation is addressed by only one question. This approach may be prone to errors. We chose to have a shorter questionnaire without repeating questions in order to prevent from early abandon of the participant and to maximize the number of complete answers. This was effectively the case because 100\% of the answers were complete. 
Furthermore, redundant question might be prone to inconsistent answers that would lead to discard them. Then, we designed the questionnaire to separate the different dimensions as much as possible to avoid ambiguity in the analysis of results.

The third limitation pertains to the relatively modest number of collected responses.  Acquiring 124 completed questionnaires spanned several months, and an increased number of participants would have necessitated alternative dissemination strategies. Nonetheless, considering the nature of the questions and the results, we deemed this sample size to be sufficient for statistically significant analysis. Notably, the substantial disparities observed in the outcomes substantiate the validity of our findings.

The quality of the collected responses is buttressed by two questions: a preliminary eliminatory question and a second question on the scope of negation, which were used to filter out participants who could bias the results. The very low number of such participants indicates that the response set is of good quality, suggesting that participants answered the questions conscientiously.

Another potential bias of this questionnaire is the presentation of basic notions of sequential patterns, which may have influenced certain responses over others. It is noteworthy that the questions on non-inclusion and embedding exhibited low diversity of responses. We expected a more varied perception of the notion of non-inclusion of itemsets, but this diversity was not reflected in the participant panel. Considering the diversity observed in the responses to the multiple occurrence question, we believe that if there was significant heterogeneity in the previous questions, it would have emerged in the questionnaire responses. Among the presentation biases, the use of symbols (rather than letters) in the questionnaire format was reported as interesting by some participants. Using letters assumes an order in the items that does not exist. In practice, we observed that only 22.6\% of participants were sensitive to the item order. The use of geometric symbols better captures the idea of set without order. Unfortunately, we did not collect information on the graphic mode of the participant used, so we cannot test this hypothesis.

Lastly, the questionnaire is closely aligns with the analysis framework proposed by Besnard and Guyet~\cite{besnard2020semantics}, which makes specific assumptions about the syntax and semantics of patterns with negation. Two crucial assumptions revolve around insensitivity to item order within an itemset and the scope of negation. The latter assumption saw 11.3\% of participants responding differently than anticipated. As we excluded these individuals from the analysis, it does not affect the conclusions, but it raises questions about the ``intuition'' held by these people. Further in-depth  interviews could shed light on this matter. A third hypothesis pertains to the syntax of patterns with negation. A more comprehensive study could explore more extensive syntaxes, such as allowing consecutive negations or negations at the beginning or end of a pattern. While such possibilities are inherent in some state-of-the-art pattern extraction algorithms, they were not explored in this study.

\section{Conclusion}
This paper delves into the semantics of sequential patterns with negation from the perspective of potential users of algorithms that extract such patterns. Prior research has highlighted the inherent ambiguity in the notations employed for these patterns~\citep{besnard2020semantics}. Our primary objective was to determine whether the patterns extracted by state-of-the-art algorithms could potentially lead to  misinterpretation by users. 

To address this question, we conducted a survey targeting potential users with diverse profiles. The goal of the survey was to understand which of the identified semantics were preferred by or intuitive for the users.

Analysis of the questionnaire responses, which involved 124 participants, revealed that two semantics dominate within the panel. A first important result is that there is no universally shared intuitive semantics among the participants.

The second significant outcome underscores the discrepancy between user intuitive semantics and the semantics used in state-of-the-art pattern extraction algorithms with negation, such as eNSP and {\sc NegPSpan}. As the partial non-inclusion arises when negation involves sets of items (e.g., $\neg(ab)$), patterns incorporating this form of constraint warrant special attention to ensure optimal user comprehension.
Furthermore, the substantial majority preference (approximately 69\% of participants) is for weak-embeddings, aligning with the choice made by the {\sc NegPSpan} algorithm. This semantics also exhibits antimonotonicity properties when negations are restricted to singletons.

Based on these findings, we offer the following recommendations for sequential pattern extraction methods with negation:
\begin{itemize}
\item Limit the use of item set negation and prioritize item negation instead,
\item Alternatively, explore the extension of the {\sc NegPSpan} algorithm, as its inclusion relation semantics aligns with the majority intuition,
\item Promote the use of specific syntaxes for each semantics in order to avoid confusion.
\end{itemize}

\bibliographystyle{plain}

\end{document}